\documentclass[12pt, a4paper]{article}
\usepackage[utf8]{inputenc}
\usepackage[pdftex]{graphicx}
\usepackage[lofdepth,lotdepth,labelformat=empty]{subfig}
\usepackage{rotating}
\usepackage{pslatex}
\usepackage{colortbl}
\usepackage[margin=1in]{geometry}
\usepackage{setspace}
\usepackage{lscape}
\usepackage{multirow}
\usepackage{ctable}
\usepackage{amsmath}
\usepackage{natbib}
\usepackage{color}
\usepackage{eurosym}
\usepackage{booktabs}
\usepackage{dcolumn}
\newcolumntype{.}{D{.}{.}{-1}}
\usepackage[toc,page]{appendix}

\usepackage{color}
\definecolor{darkred}{rgb}{0.5,0,0}
\definecolor{darkgreen}{rgb}{0,0.5,0}
\definecolor{darkblue}{rgb}{0,0,0.5}

\usepackage{hyperref}
\hypersetup{colorlinks,
 linkcolor=black,
 filecolor=darkgreen,
 urlcolor=darkblue,
 citecolor=darkblue,
}

\begin{document}

\title{Detecting Policy Preferences and Dynamics in the UN General Debate with Neural Word Embeddings\thanks{UN General Debate Corpus is available on the Harvard Dataverse at \url{http://dx.doi.org/10.7910/DVN/0TJX8Y}.}}

\vspace{-0.5in}
\author{
Stefano Gurciullo\\
  University College London\\
  \href{mailto:stefano.gurciullo.11@ucl.ac.uk}{stefano.gurciullo.11@ucl.ac.uk}
\and
Slava J. Mikhaylov \\
 University of Essex\\
  \href{mailto:s.mikhaylov@essex.ac.uk}{s.mikhaylov@essex.ac.uk}\\  
 }



\maketitle

\begin{abstract}
\footnotesize
\noindent Foreign policy analysis has been struggling to find ways to measure policy preferences and paradigm shifts in international political systems. This paper presents a novel, potential solution to this challenge, through the application of a neural word embedding (Word2vec) model on a dataset featuring speeches by heads of state or government in the United Nations General Debate. The paper provides three key contributions based on the output of the Word2vec model. First, it presents a set of policy attention indices, synthesizing the semantic proximity of political speeches to specific policy themes. Second, it introduces country-specific semantic centrality indices, based on topological analyses of countries' semantic positions with respect to each other. Third, it tests the hypothesis that there exists a statistical relation between the semantic content of political speeches and UN voting behavior, falsifying it and suggesting that political speeches contain information of different nature then the one behind voting outcomes. The paper concludes with a discussion of the practical use of its results and consequences for foreign policy analysis, public accountability, and transparency. 
\end{abstract}

\thispagestyle{empty}


\section{Introduction}
Political systems appear to be different from several other complex systems, for the central role that ideas -- be they world views, norms, or policy programs -- feature in the shaping of political actors' behavior \citep{campbell2002ideas}. The challenge is, though, to find a consistent method to quantitatively measure political agents' positions with respect to such ideas. Qualitative approaches traditionally dominated this area through case studies or historical analyses \citep{mansbachrichard2004revolutions,meyer2013conference}. While the contribution of this line of scholarship is of great value, it lacks the reproducibility and generalizability that quantitative methods claim.

Foreign policy researchers have started considering voting behavior as a proxy for policy preferences towards various policy themes or ideas. In the case of the United Nations, Voeten \citep{voeten2012data} extracts ideological positions of UN member states, by measuring their relative voting difference with the voting patterns expressed by the United States. The attempt represents, indeed, a valuable step forward, but it lacks in so far as it is not possible to extract the position of countries with respect to specific policy ideas. Voting is the final, strategic outcome of a constrained process on specific issues contained in UN Resolutions, not necessarily embodying countries' attitudes towards greater scale policy themes \citep{schwarz2014estimating}. 

In contrast, political speeches and texts, with their less constrained and potentially higher informational content, may well be more suited to provide a source of measurement. This paper contributes to this line of work by introducing the use of neural word embedding (Word2vec) models on political text -- more specifically, UN General Debate (UNGD) speeches. It is postulated that Word2vec extracts what can be considered as the semantic position of single words and documents in vectorspace, thereby allowing to quantify how semantically 'close' speeches by single or more nations are to specific concepts or to other nations. 

Based on this interpretation of Word2vec, the paper proposes an evaluation of two types of semantic indices, one that monitors nations' closeness to pre-determined policy concepts, and one that detects countries' relative semantic position with respect to each other. The remainder of the paper is structured into five sections. Section 2 discusses related work, and provides a concise account of political science scholarly efforts to extract quantitative insights from text, as well as applications of Word2vec that are adjacent to the paper's focus of attention. Section 3 introduces the data, which comprises UN General Debate speeches from 1970 to 2014. Section 4 explains the methodology of this study, covering all stages from data pre-processing to the indices' calculations. Section 5 shows the results, and provides an interpretation of the selected indices. It also tests whether there exists a statistical relation between the model's results and UN voting patterns. A concluding section follows, briefly discussing potential usages of the semantic indices, as well as their limitations and future work.

\section{Related Work}

The use of natural language processing in detecting political preferences is not, by itself, new, nor is Word2vec's application in social systems. This section very briefly reviews these two strands of literature, on which this paper draws. 

\textbf{Political preferences and quantitative text analysis}. Political science has routinely drawn on natural language processing and machine learning to extract political actors' behavior. As for supervised learning, early applications include the Wordscore model \citep{laver2003extracting}, a Naive Bayes classifier that categorizes text across a one-dimensional spectrum, where the ends of the spectrum are learnt by training on the bag-of-word features of a set of anchor corpora. The model has been used to track the ideological differences between US Republicans and Democrats \citep{kidd2008real}, detect conflict in political preferences \citep{monroe2008fightin}, and interest group influence \citep{kluver2009measuring}. Other supervised methods, such as random forests and support vector machines, have been applied to classify political text into one or more ideological dimensions. Applications have included Stewart and Zhukov \citep{stewart2009use}, which used decision tree ensembles to monitor Russia's defense policy during the Georgia-South Ossetia conflict. A complete review and evaluation of supervised learning techniques applications on political text is provided by Grimmer \citep{grimmer2013text}. 

Unsupervised learning approaches have been of special interest to political scientists, as they can reveal useful information in extracting the topics of interest to political actors without the input from a human observer. One line of work has been dedicated to reduce the dimensionality of political text, and identify to what policy themes the reduced dimensions correlate to. A standard dimensionality reduction technique used in the literature is Correspondence Analysis (CA), which is a version of the more widely known in computer science literature Latent Semantic Indexing (LSI). CA is traditionally considered better suited for discrete-state features, such as text, than Principal Component Analysis \citep{greenacre2007correspondence}. It has been applied, for instance, to measure the political preferences of candidates in state and federal US elections, by testing the existence of a relation between the reduced dimensions and voting data \citep{bonica2013ideology,bonica2014mapping}. Other studies have applied topic models, most notably Latent Dirichlet Allocation \citep{blei2003latent} to infer the policy topics that characterize a set of political documents. Applications have included \cite{gurciullo2015complex}, which adopted dynamic topic modeling \citep{blei2006dynamic} to infer and monitor the salience of policy themes across individuals and parties in the UK House of Commons. 

\textbf{Neural word embeddings, Word2vec}. Word2vec  is a two-layer neural network model developed by researchers led by Tomas Mikolov at Google, Inc. \citep{mikolov2013efficient}\footnote{http://deeplearning4j.org/word2vec}. It creates vectors that are numerical representations of words' features, where the features are their context - i.e. the set of words or characters which tend to appear within a certain range of that word. The result is a framework that permits -- given enough data as a training sample -- to capture the semantics of words, by observing how they group in the high-dimensional vectorspace. A complete technical overview of Word2vec's learning algorithm is not within the scope of this paper. \cite{goldberg2014word2vec} and \cite{rong2014word2vec} provide excellent accounts.

Word2vec models are slowly but steadily finding room for application in the social sciences. In economics, they have been used to develop a novel method of technology forecasting \citep{nikitinsky2015introducing}. More specifically, they have been applied to documents about governments' industrial contracts, identifying technology areas which appear to be of greater interest for public institutions, and therefore more likely to experience more funding and faster development. Finance scholars have used neural word embeddings on documents from the Multiple Listing Service (MLS), the most important real estate database in the United States, in order to perform property ranking and assess their investability \citep{shahbazi2016estimation}. They do so by measuring the similarity of documents related to any one property to a pre-determined set of words that defines good estate conditions. Word2vec has also been used in marketing and pricing, as shown in \cite{lv2015network}. Based on user comments data from e-commerce websites, the scholars use neural word embeddings to extract indices of consumers' perceived value of a range of products. The output of these efforts would then better inform the pricing strategy of online retailers.  

So far, neural word embeddings are registering impact on the economic sciences, and not - to the authors' knowledge - on political science and more specifically on the tracking of political preferences. The paper thus intends to lay down the first steps for the successful implementation of such machine learning models in this area of study.

\section{Data}
This paper's dataset comprises all speeches made at the General Debate of the United Nations General Assembly (UNGA), from 1970 to 2014. The General Assembly is the main deliberative body of this international organization, comprising all member nations at a given point in time\footnote{http://www.un.org/en/ga/about/}. The General Debate is an annual activity that opens new UNGA session. Taking place over several days, the General Debate schedules speeches by one representative from each UN member state. Most of statements are delivered by heads of state and government, with a smaller proportion delivered by foreign ministers, and a minority of statements delivered by heads of country delegations at the UN. These speeches are used by countries to express their views on current issues in foreign affairs, and are regarded as an invaluable source of information to understand their policy preferences -- a sort of ``barometer of international opinion'', as stated by Smith \citep{smith2006politics}.

This project utilizes UN General Debates corpus that was first introduced in \cite{baturo2016general}.\footnote{\normalsize The UNGDC is publicly available on the Harvard Dataverse at \url{http://dx.doi.org/10.7910/DVN/0TJX8Y}. A browsing tool that allows users to explore individual documents and the topics covered, including the top words that characterize topics, the evolution of topics over time, word distributions across topics, the underlying digitized texts of speeches, and the source PDFs is available at \url{http://ungd.smikhaylov.net}.} The corpus contains all country statements made during General Debate retrieved through the UN Bibliographic Information System. All speeches are available in six official languages, and the UN General Debates corpus contains official English language versions of speeches.  There is only one speaker per country in the General Debate. It follows that the unit of analysis is a country-year. 

\begin{figure}[h!]
\centering
\includegraphics[width = \textwidth]{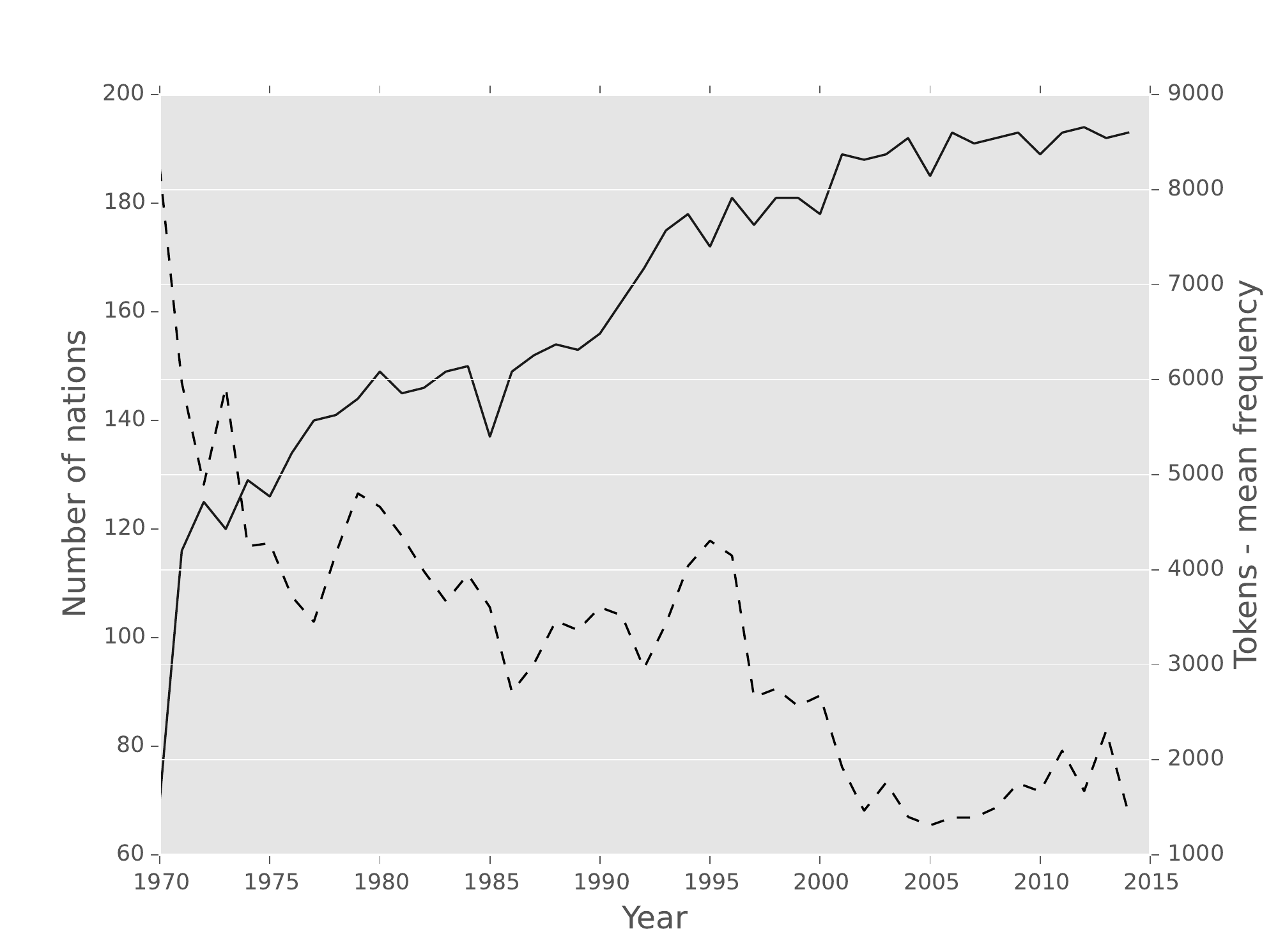}
\caption{Number of nations and mean token frequency per document, 1970-2014.}
\label{fig:graph}
\end{figure}

The corpus is composed of 7,310 documents (one per country-year), with an average of 945 unique tokens per country-year. Figure \ref{fig:graph} shows the number of country-years and mean token frequency over the entire time period considered. As the number of UN member states grows, speeches gets shorter. The UN General Debates corpus is available in supplementary materials. 

\section{Methodologies}
The methods behind the analysis involve three steps. The first is text pre-processing, which is followed by the implementation of a variant of the Word2vec model that turns into vectorspace also labeled sets of words, known as Doc2vec. Doc2vec's output is then used to (1) evaluate an index of semantic similarity between nations and policy themes, (2) evaluate two indices of semantic centrality in a network of nations, (3) test the existence of a relation between UN voting behavior and the Doc2vec vectorspace.

\subsection{Corpus Preprocessing} 
The corpus is grouped into documents, each enclosing speeches made by country representatives in a UN General Debate session. All documents in the corpus are stripped of digits and all text is lower-cased. Because of the importance of contextuality as a feature for the Word2vec model, it is deliberatively chosen not to remove punctuation, nor stopwords. The documents are then tokenized, and a snowball stemmer \citep{porter2001snowball} is implemented in order to reduce similar words to their root form (e.g. the tokens 'economy' and 'economic' would reduce to `econ'). All words with total frequency less than 5 are discarded, resulting in 16,765 unique words, plus 7,310 labeled documents, for a total of 24,075-element vocabulary for the learning model. Preprocessing is performed using the Python NLTK package\footnote{http://www.nltk.org/}. 

\subsection{The Learning  Model}
Word2vec has the task of predicting a word given the other words in a context, and it does do by representing words as n-dimensional vectors, where the number of dimensions is defined by the user. Words that have more proximity in this n-dimensional space appear to also have semantic closeness, given the raw data fed into the model. 

As stated in \cite{mikolov2013efficient}, more formally, given a sequence of training words $w_1, w_2, w_3, ..., w_T$, the task is to maximize the average log probability:

\begin{equation}
\frac{1}{T} \sum_{t=k}^{t-k} \log p(w_t|w_{t-k},...,w_{t+k})
\end{equation}

In the neural network implemented by this paper, the prediction task is done via hierarchical softmax \citep{morin2005hierarchical}.

This analysis makes use of a variant of Word2vec, generally known as Doc2vec \citep{le2014distributed}, that treats compounds of words as single elements, thereby contributing to the prediction task set above. In this paper's case, the compounds are the 7,310 country-year documents. The learning process is an extension of the more conventional Word2vec. More detailed discussion is outside the scope of this paper, but covered in the above references. 

The authors implement the Doc2vec model in Python through the Gensim library\footnote{https://radimrehurek.com/gensim/}. Two key inputs affect the outcome of the model: the size of the vectors and the window, i.e the maximum distance between the predicted word and the words used for prediction within a document. After several trials, it has been opted to set the vector size to 200, and the window to 10. The initial learning rate is set to 0.025.

\subsection{The Semantic Indices}
The output of the Doc2vec model is a matrix of the size of the vocabulary times the size of the vectors. In order to construct the semantic indices, a measure of similarity between vectors is needed. In line with much of the literature on applications of neural word embeddings, it is opted for cosine similarity. Given token $a$ and $b$, their cosine is simply the dot product of their vectors, divided by the product of the vectors' norms:

\begin{equation}
C(a,b) = \frac{\vec{a} \cdot \vec{b}}{|| \vec{a} || \ ||\vec{b}||}
\end{equation}

\textbf{Topic-related semantic index}. A topic-related semantic index is defined as the deviation in the arithmetic mean cosine similarity of a group of country-years (or simply the cosine similarity, in case of only one country) and a set of tokens deemed to represent a policy theme, with respect to a base year. 

The index is best explained through an example. Say, it is intended to calculate the topic-related index for country group $N = [n_1,n_2,n_3,...,n_n]$, with respect to a policy theme represented by set of tokens $W = [w_1, w_2, w_3,...,w_k]$. Let year $y_b$ be the base year, and year $y_t$ the year of interest. The topic-related semantic index $I^{N,W}_{y_t}$ is evaluated by:

\begin{equation}
I^{N,W}_{y_t} = \frac{\frac{1}{n} \sum_{i=1}^{n} C_{y_t}(n_i,W) - \frac{1}{n} \sum_{i=1}^{n} C_{y_b}(n_i,W)}{|\frac{1}{n} \sum_{i=1}^{n} C_{y_b}(n_i,W)|}
\end{equation}

Where $C_{y_t}(n_i,W)$ is the cosine similarity between nation $i$ and words $W$ at year ${y_t}$. Note that the cosine similarity with a group of words can be easily performed by taking into account the mean across their respective vector values. $I^{N,W}$ is real-valued, approaching zero when the mean similarity at a given point in time is close to the base year. Negative values imply a semantic similarity that is less than in the base year, vice versa when positive values are observed.

This study evaluates topic-related indices for the entire UNGA membership, with the aim of identifying macro-scale changes in policy preferences. Obviously, any other arbitrary choice of countries is possible. Four indices are calculated, each referring to a different policy theme, operationalized by a couple of key words, as shown in Table \ref{tab:Table 1}. 

\begin{table}[htbp]
  \centering
  
    \begin{tabular}{rr}
    \toprule
    \textbf{Policy theme} & \textbf{Key words} \\
    \midrule
    Health & `health', `sanit' \\
    Education & `educ', `school' \\
    Nuclear weapons & `nuclear', `weapon' \\
    Islamic terrorism & `terror', `islam' \\
    \bottomrule
    \end{tabular}%
   \caption{Policy themes considered and respective key words}
  \label{tab:Table 1}%
\end{table}%

\textbf{Semantic centrality indices.} The results from the neural word embedding model allow to explore how similar countries' texts are, where the similarity is meant both with regards to the topics contained in the document and to the style. It is therefore feasible to build a topological analysis based on countries' similarities. By inspecting the structure of such networks, it is possible to measure the change in the centrality of nations. 

At each year, an adjacency matrix $M_t$ that represents the semantic network is constructed, by taking the cosine similarity as per equation (2) for each pair of countries. The result is a weighted complete graph, where the weights are the cosines. $M_t$ is then filtered. Edge $C_t(a,b)$ in $M_t$ is removed if it does not satisfy both conditions below:
\begin{enumerate}
  \item $C_t(a,b)$ must be above the 95th percentile of country $a$'s and/or country $b$'s edge weight distribution. 
  \item $C_t(a,b)$  must be greater than 0.6\footnote{This arbitrary value has been chosen after several trials, and considered to provide sensible but balanced filtering.}.
\end{enumerate}

After the filtering, the graph becomes sparse, retaining what are deemed to be the most significant relations. 

Two semantic centrality indices are created. The first is the deviation from the mean of the eigenvector weighted centrality for any given country or group of countries. Let $\epsilon^i_t$ be the weighted eigenvector centrality for country $i$ at year $t$, evaluated as in \cite{bonacich1987power}. Index $E^i_t$ is equal to: 

\begin{equation}
E^i_t = \frac{\epsilon^i_t - \frac{1}{n}\sum_{j=1}^{n}\epsilon^j_t}{\frac{1}{n}\sum_{j=1}^{n}\epsilon^j_t}
\end{equation}

In order to obtain index $E$ for a group of countries, it suffices to substitute $\epsilon^i_t$ in equation (4) for the mean of the eigenvector centralities of the countries in the group. This paper exemplifies the use of $E$ by evaluating and comparing it for two groups of countries, the EU-15\footnote{The first 15 countries who joined the European Union, namely Austria, Belgium, Denmark, Finland, France, Germany, Greece, Ireland, Italy, Luxembourg, Netherlands, Portugal, Spain, Sweden, United Kingdom.} and a set of emerging economies\footnote{Brazil, China, India, Indonesia, Mexico, Russia, South Africa, South Korea, Turkey, Saudi Arabia.}.

For a clearer understanding of how $E$ changes over time, it is possible to measure its deviation from a base year $y_b$. Let $\dot{E^i_{y_t}}$, the second index, be defined as:

\begin{equation}
\dot{E^i_{y_t}} = \frac{E^i_{y_t} - E^i_{y_b}}{|E^i_{y_b}|}
\end{equation}

In section 5, the index $\dot{E^i_{y_t}}$ is used to analyze the relative semantic importance of Russia and the United States of America over the entire time period considered.

\begin{figure}[ht!]
            \includegraphics[width = 0.5\textwidth]{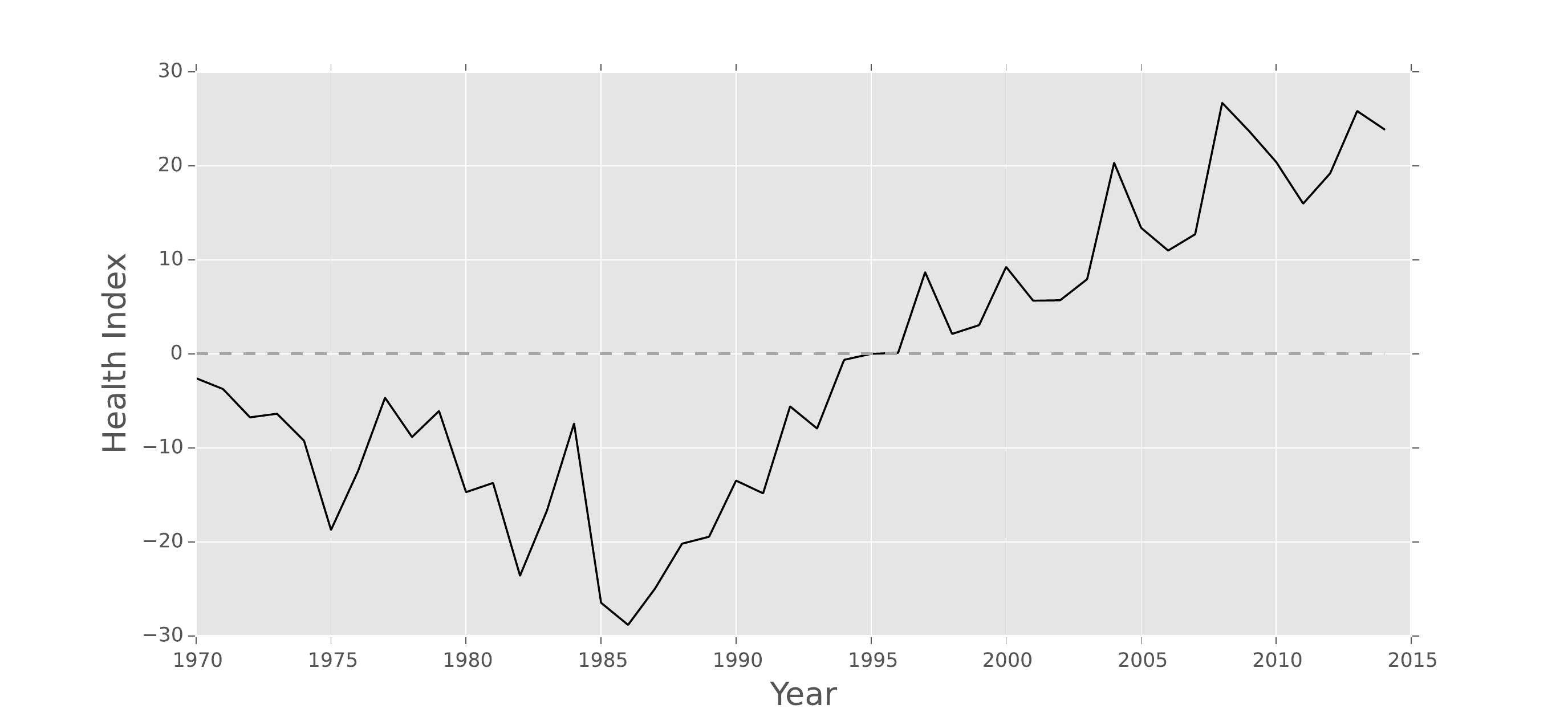}
           \includegraphics[width= 0.5\textwidth]{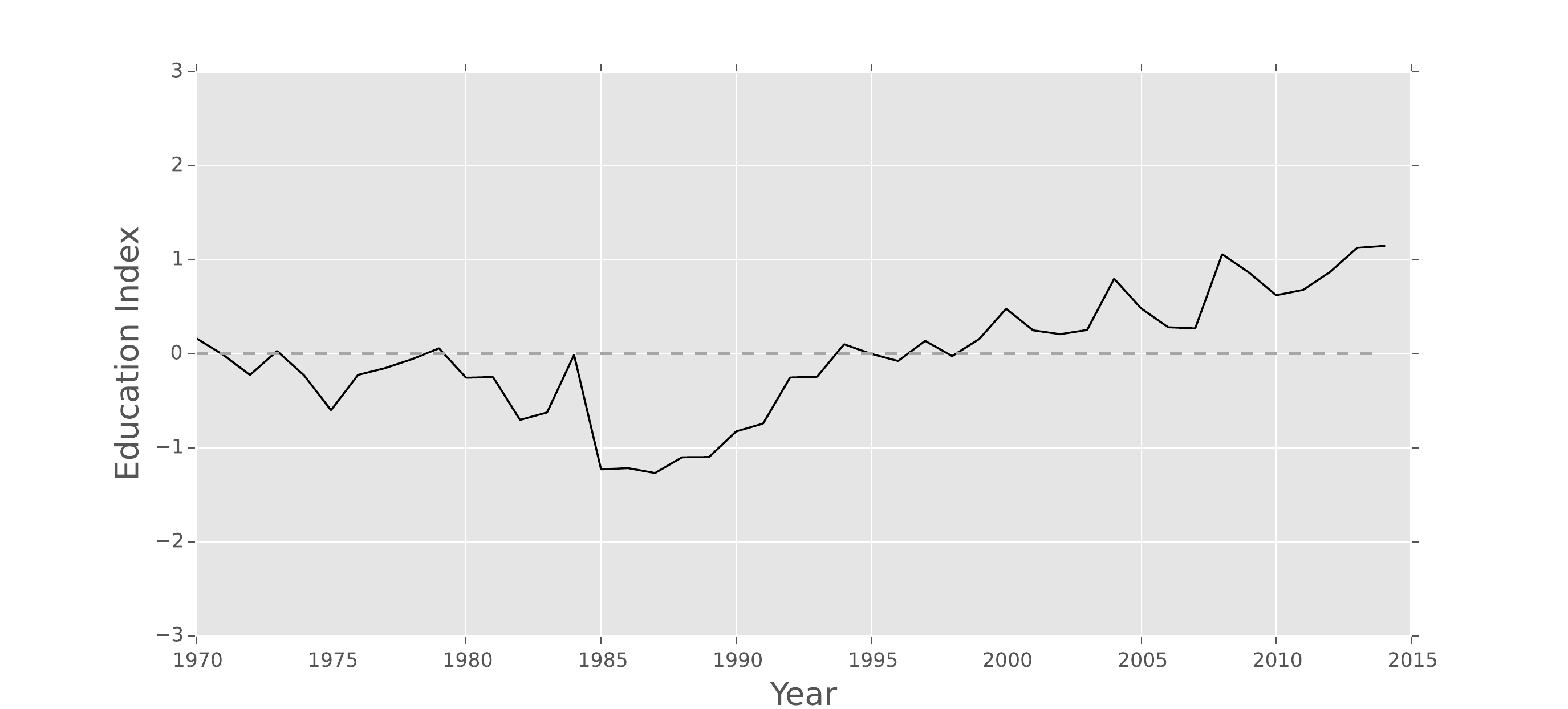}\\
            \includegraphics[width=0.5\textwidth]{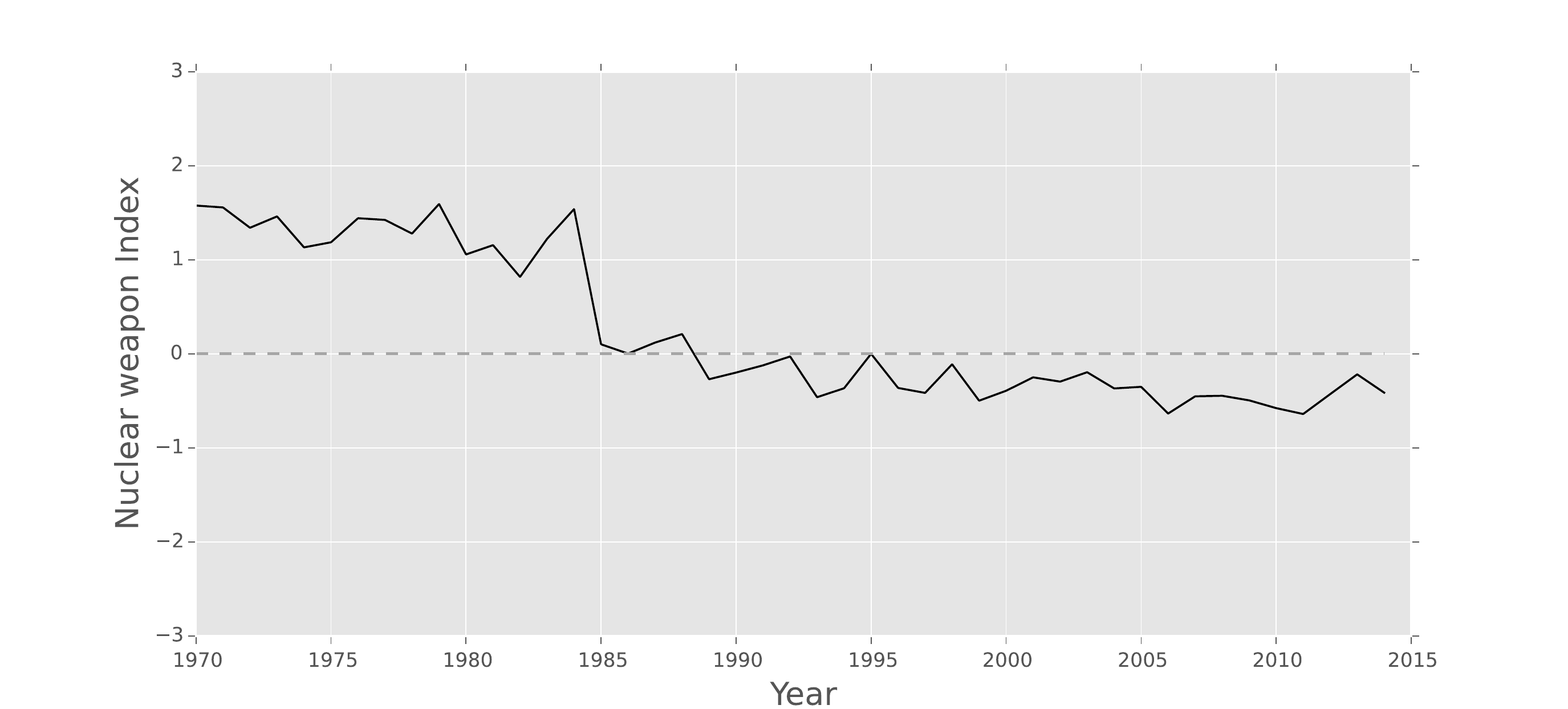}
            \includegraphics[width=0.5\textwidth]{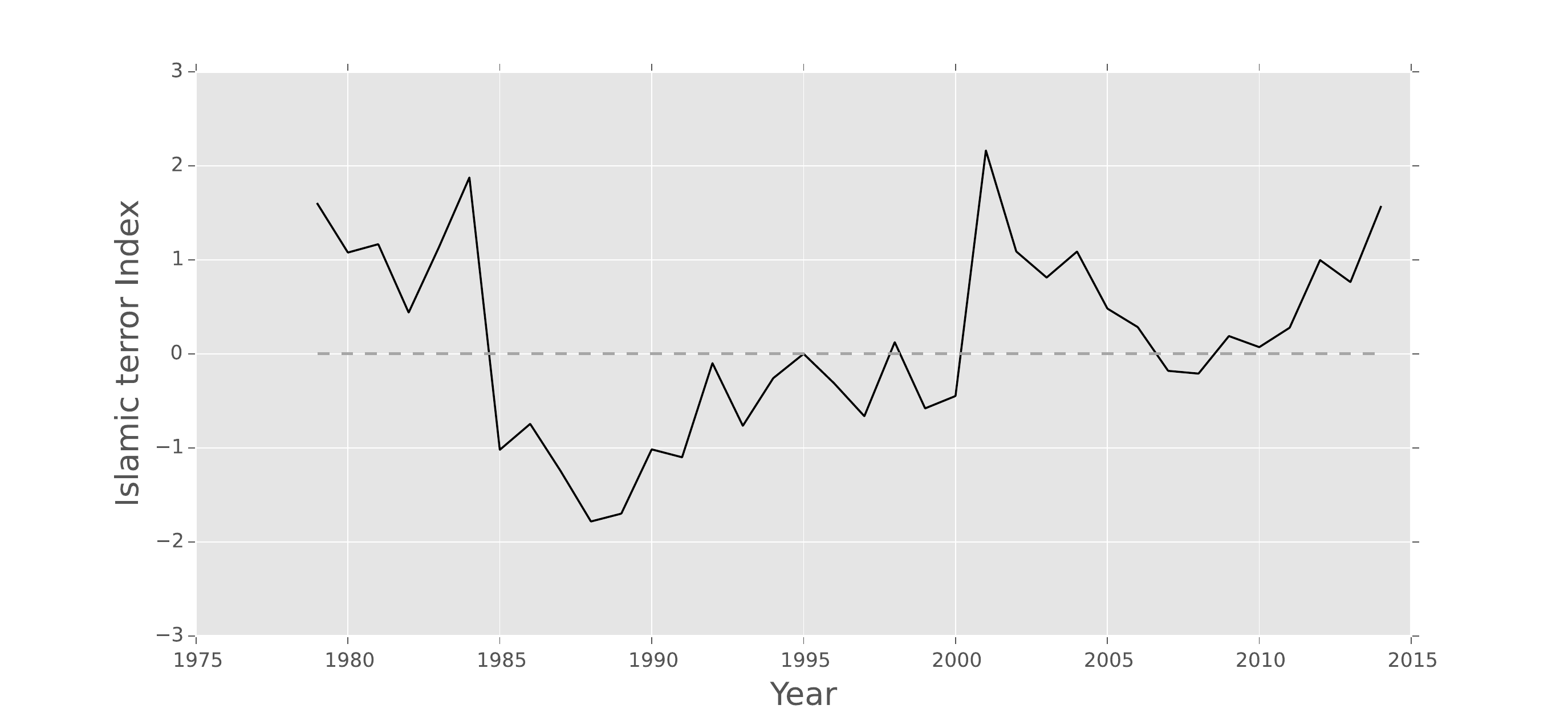}  
            
 \caption{Topic-related semantic indices for education, health, nuclear weapons and Islamic terrorism, with base year = 1995.}
   \label{fig:figures}
\end{figure}

\textbf{Testing a relation between neural word embeddings and voting behavior}. As a final exercise, the existence of a statistical relationship between the model's results and UNGA voting behavior is tested. This is performed by inspecting Spearman's correlation coefficient \citep{zar1972significance} between two vectors: $\vec{C_{us}}$, containing all cosine similarities between USA speeches and all other countries' speeches for each year, and $\vec{V_{us}}$, recording the percentage of each country's votes that matched USA's votes for each year. The latter data is provided by \cite{voeten2012data}\footnote{Available at http://bit.ly/1PuAX2O}. Further testing and insights are then provided in the supplementary information.

\section{Results}

\textbf{Exploring UNGD's semantic closeness to selected policy themes}. Figure 2 shows the dynamics of the topic-related semantic indices with regards to the four policy themes selected. They summarize the average semantic similarity of the speeches of all members of the UN to the key words that represent such topics. 

Their dynamics appear to be well explained by historical events. Both the education and health indices feature lower levels until about 1985, the year Mikhail Gorbachev came to power in the USSR. With Gorbachev and the subsequent end of the Cold War, policies at the United Nations started becoming less focused on security issues and more on social and development goals \citep{simes1987gorbachev}. The steady, increasing trend of the indices since then can be explained by more and more focus on the Millennium Development Goals policy efforts, which retain both education and health at their core \citep{sachs2005millennium,griggs2013policy}.

In a similar vein, the nuclear weapons index experiences an abrupt decline with Gorbachev's rise to power, featuring a slow declining trend since then, thus suggesting that the topic has been having much less importance than during nuclear arms race. The Islamic terror index is more volatile and responsive to large terrorist attacks. Starting in 1979 -- the year experts regard as the commencing point for organized fundamentalist Islamic terror \citep{o1997islamic} -- the index registers high levels, in concomitance with the bombings of US embassies in Lebanon and Kuwait \citep{quillen2002historical}. A new spike occurs in 2001, most likely due to the 9/11 attacks. A rising trend is observed during the last years of the index, explained by the rising threat from ISIL, who has targeted both civilians and UNESCO's sites \citep{harmancsah2015isis}.

\begin{figure}[h!]
\centering
\includegraphics[width = \textwidth]{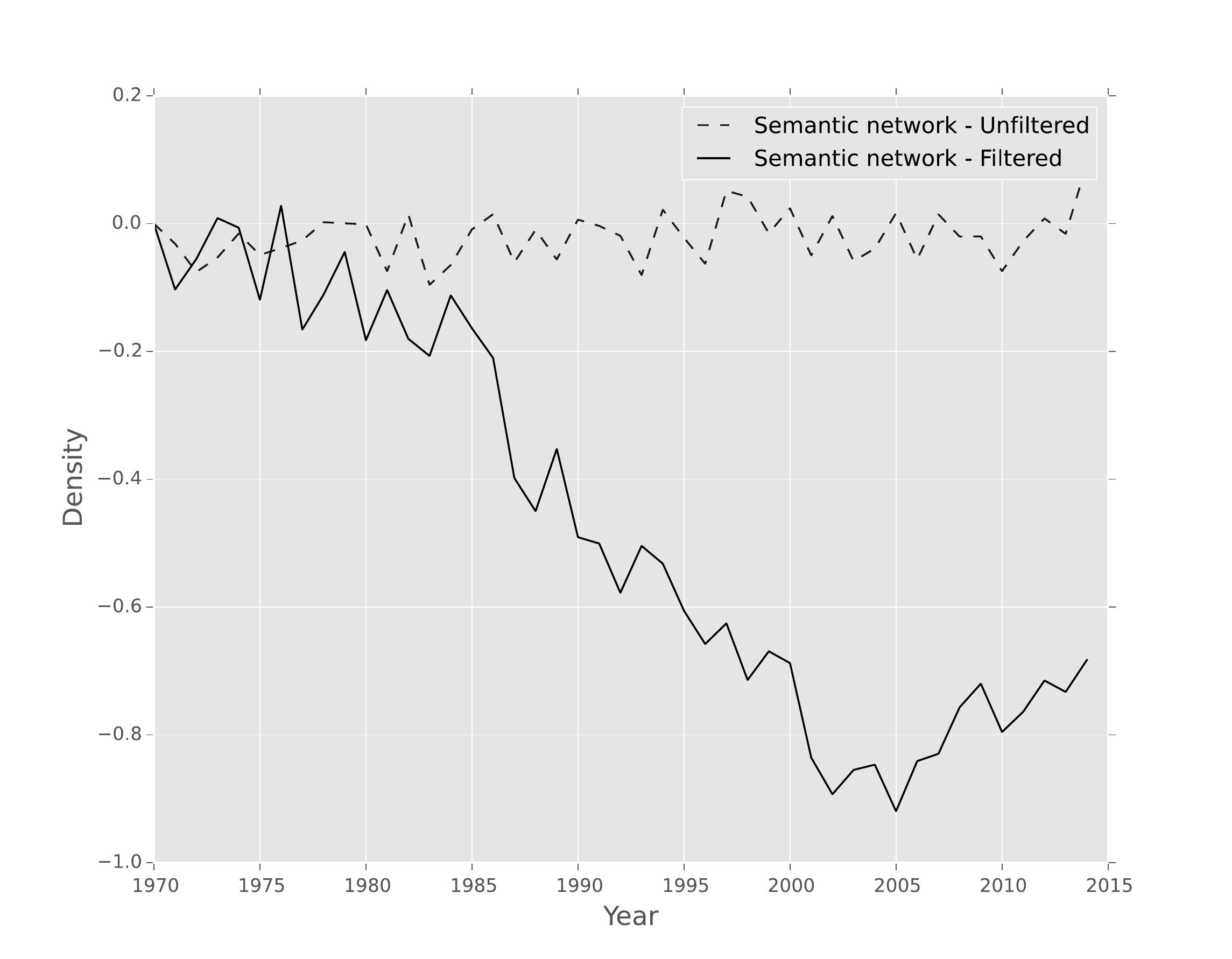}
\caption{Density index of the semantic network, unfiltered and filtered form, 1970-2014, base year = 1970.}
\label{fig:digraph}
\end{figure}

\textbf{Exploring the semantic centrality for a selected group of countries}. As explained in section 4.3, for each year it is built a semantic weighted undirected graph where edges are weighted by the cosine similarity between countries. Figure \ref{fig:digraph} depicts how its density changes, with base year 1970. The dashed line refers to a semantic network with only  filter (1) applied (see Methodologies). Density appears to be stationary, with not very different levels from year to year. The situation changes when both filters are applied, with a plunge in density during the rise of Gorbachev and the end of the Cold War. The trend appears to indicate that UNGA members, as their number grew and a radical paradigm shift in the world order occurs, are less semantically close. A more diversified set of topics and style positions them farther away according to the similarity metric used. 

\begin{figure}[h!]
\centering
\includegraphics[width = \textwidth]{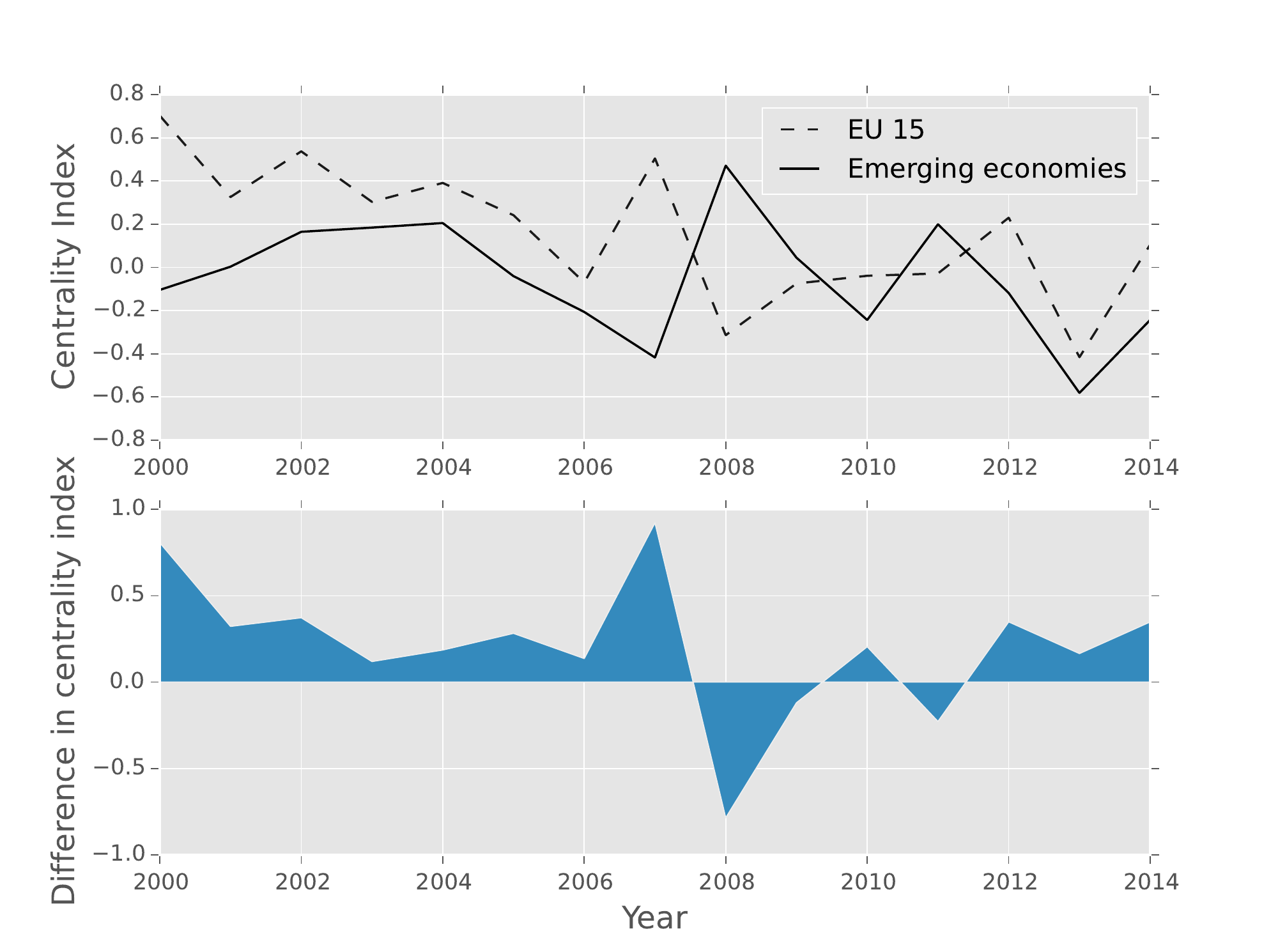}
\caption{Semantic centrality index $E$ for the EU-15 and emerging economies, 2000-2014.}
\label{fig:first}
\end{figure}

Interesting patterns are found also by the centrality indices. Figure \ref{fig:first} compares centrality index $E$ for the EU-15 and a set of emerging economies between 2000-2014. European nations appear to be more central in the semantic network until 2008, when the financial crisis hit several Western economies. After that, the difference in the indices is less stable. A plausible interpretation for the change in trend is the increasing economic and political importance of emerging economies, which in their speeches confront policy topics that gain relatively more attention by other UNGA members.

\begin{figure}[h!]
\centering
\includegraphics[width = \textwidth]{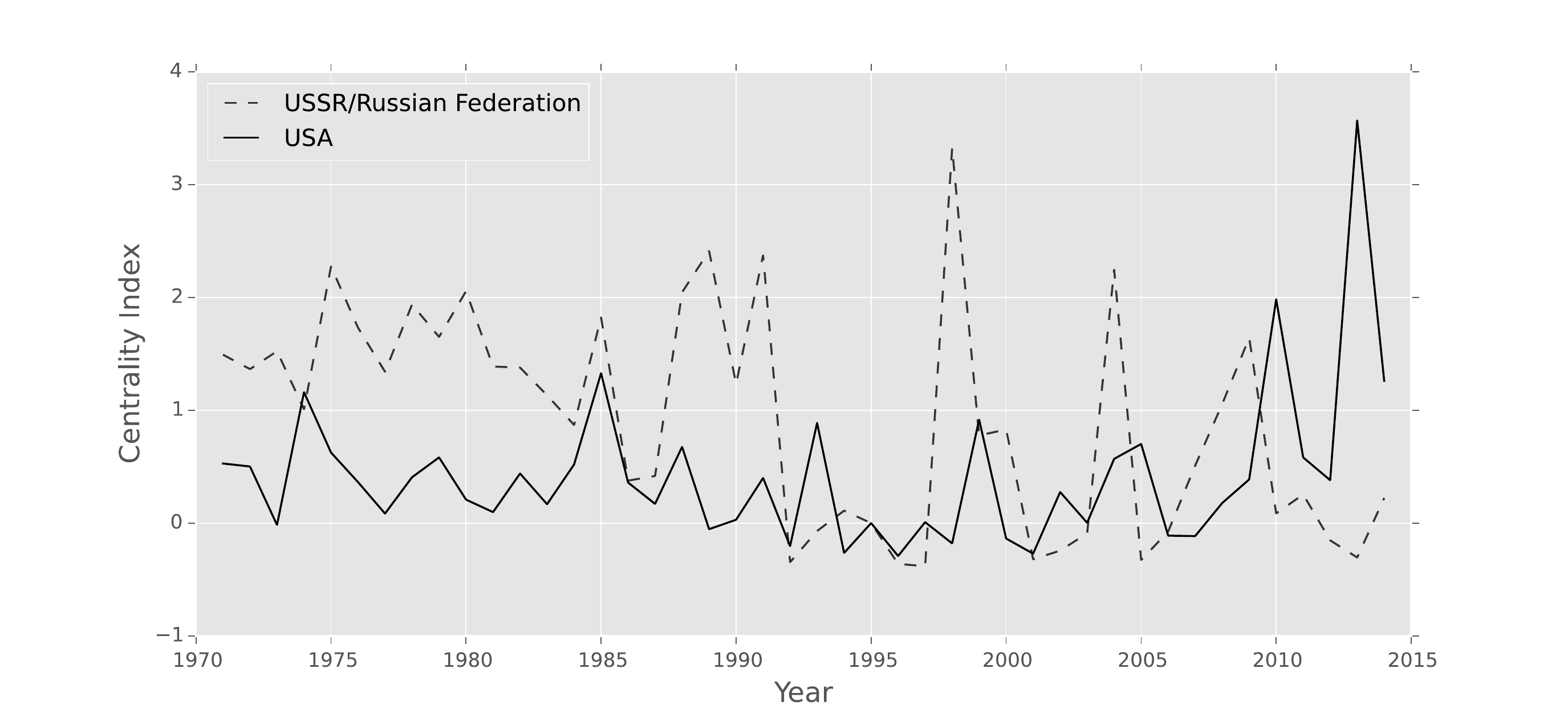}
\caption{Semantic centrality index $\dot{E}$ for USA and Russia, 1970-2014, with base year = 1995.}
\label{fig:third}
\end{figure}

Figure \ref{fig:third} compares semantic centrality index $\dot{E}$ for the United States and Russia, identified as the USSR before its dissolution in 1991. While US' index dynamics appear less volatile, apart for peaks in the late 2000s, the Russian one is extremely so. Interestingly, its extreme changes occur at the onset of important socio-economic events for the country. The steep decrease in 1991 signals the official termination of the USSR. The peaks in 1998, 2004 and 2009 co-occur respectively with the Russian financial crisis, the re-election of Putin as President, and the Russian Great Recession and gas dispute with Ukraine.

\begin{figure}[h!]
\centering
\includegraphics[width = \textwidth]{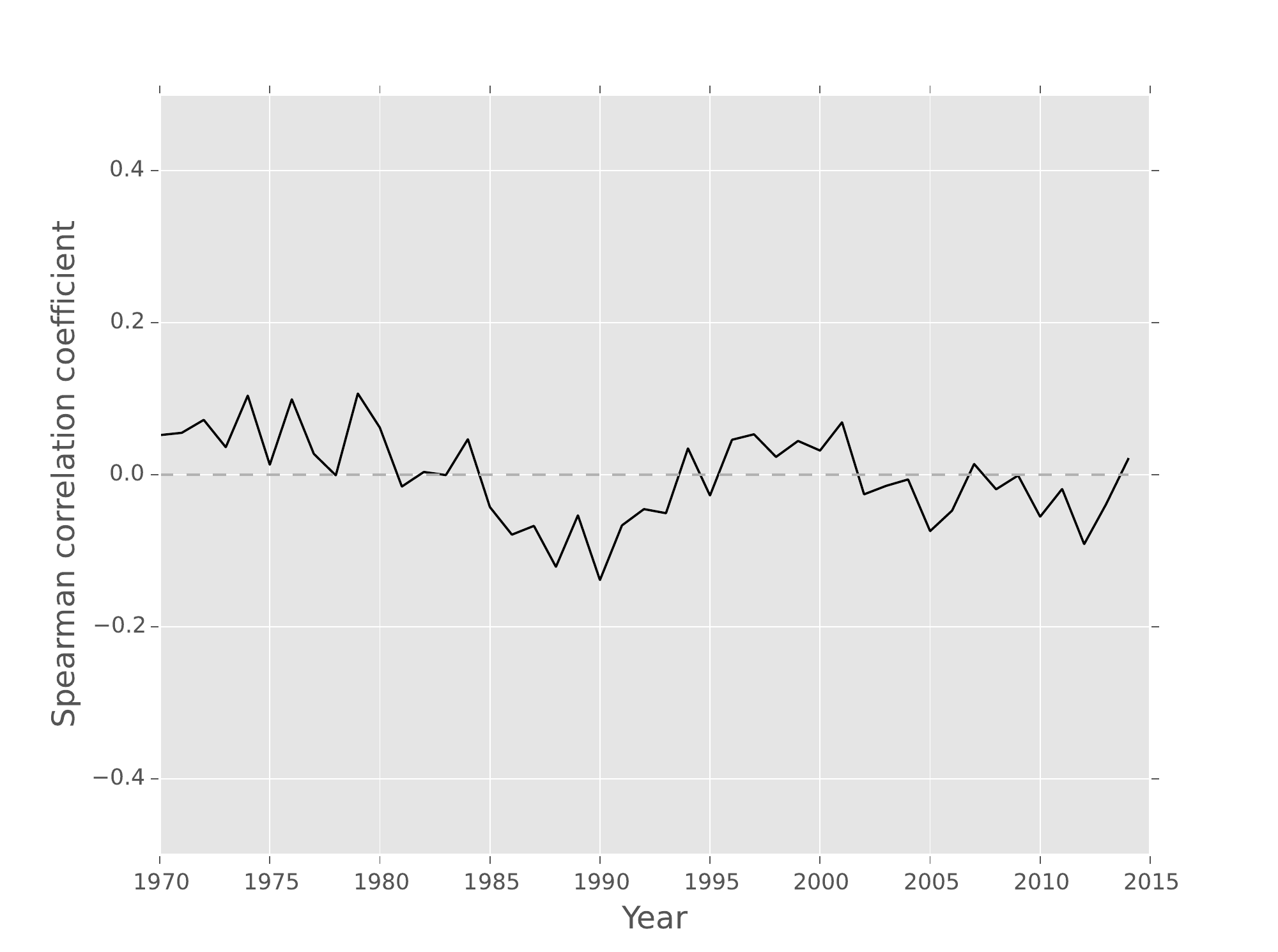}
\caption{Spearman correlation coefficient between cosine similarities of countries against USA, and countries' voting that matches USA's voting, 1970-2014.}
\label{fig:spearman}
\end{figure}

\textbf{Weak relation between model's results and voting data}. Spearman's correlation coefficient has been used as a simple test for a nonlinear relation between voting data and the results of Word2vec. The coefficient tends to be weak and close to zero most of the years considered, as shown in Figure \ref{fig:spearman}. During the Gorbachev's years, a tenuous yet stronger negative relationship seems to appear, suggesting that countries that were semantically close to the USA were not voting in a fashion similar to it. Nevertheless, the relation is too weak or absent, falsifying the contention that political textual data can yield insight into voting behavior. The two things, at least in the UN case, may well be governed by different types of factors. UN voting is the result of a strategic process happening often in informal political contexts, and on very specific policy issues; speeches represent the official point of view of countries on a wide variety of policy ideas and themes. 

\section{Concluding Remarks}
This paper has advanced the notion that neural word embeddings can be used to extract insights onto what policy themes and ideas are more salient for a nation or a group of nations, and how their textual patterns in relation to the rest of the system can reveal information about them. By applying a version of the Word2vec model, word vectors have been used to create topic-related semantic indices and graph-based semantic centrality indices. Illustration of the framework have shown the increasing salience of policy topics such as education and health, in contrast to others, such as nuclear weaponry policy, or Islamic fundamentalist terror. Graph-based insights appear to succeed in identifying paradigm policy shifts, with respect to, for instance, financial crises and political regime changes, such as the end of the Cold War. 

This strand of applications of neural word embeddings is at its infancy, and needs to be further evaluated, tested and explored. It still has to be properly understood what underlying phenomena determine the indices' dynamics. In this paper, they included a mix of both political and economic factors. Further research ought to provide a method to identify such factors without or with little human interpretation. Other efforts should concentrate on testing relationships between the indices and other processes, be they economic, financial, or political structured data. 

Despite being at its initial stages, the framework presented here has significant potential for applications in government, transparency and accountability. By monitoring topic-related indices, for instance, interest groups have the opportunity to closely observe to what extent their policy preferences are mirrored by political actors. Foreign policy analysts can obtain systematic information about both the policy interests and the semantic centrality of any determined country for which a corpus is available. The framework can be easily extended to corpora featuring higher frequencies - e.g. daily newspaper articles, monthly magazine contributions - in order to obtain semantic indices at much more granular levels.

\clearpage
\bibliography{unembeddings}
\bibliographystyle{apsr}

\end{document}